\documentclass[sigconf,natbib=true,anonymous=False]{acmart}
\usepackage[most]{tcolorbox}
\usepackage{multirow}
\usepackage{array}
\usepackage{caption}
\usepackage{subcaption}
\usepackage{graphicx}
\usepackage{colortbl}
\usepackage{xcolor}
\usepackage{subcaption}
\usepackage{amsfonts}
\usepackage{graphicx}
\usepackage{makecell}
\usepackage{subcaption}
\usepackage{booktabs}
\usepackage{arydshln}
\usepackage[table]{xcolor}
\usepackage{enumitem}
\definecolor{tableheaderblue}{RGB}{220, 230, 240}

\setcopyright{none}

\AtBeginDocument{%
  }

\setcopyright{acmlicensed}
\copyrightyear{2018}
\acmYear{2018}
\acmDOI{XXXXXXX.XXXXXXX}
\acmConference[Conference acronym 'XX]{Make sure to enter the correct
  conference title from your rights confirmation email}{June 03--05,
  2018}{Woodstock, NY}
\acmISBN{978-1-4503-XXXX-X/2018/06}

\begin{document}

\title{Retrieval-Augmented Multimodal Model for Fake News Detection}

\author{Yiheng Li}
\authornote{Both authors contributed equally to this research.}
\email{202501410213@uibe.edu.cn}
\affiliation{%
  \institution{University of International Business and Economics}
  \city{Beijing}
  \country{China}
}

\author{Weihai Lu}
\authornotemark[1] 
\authornote{Corresponding author.}
\email{luweihai@pku.edu.cn}
\affiliation{%
  \institution{Peking University}
  \city{Beijing}
  \country{China}
}

\author{Hanyi Yu}
\affiliation{%
  \institution{University of Southern California}
  \city{Los Angeles}
  \state{CA}
  \country{USA}}
\email{hanyiyu@usc.edu}

\author{Yue Wang}
\affiliation{%
  \institution{Upstart Holdings, Inc.}
  \city{San Mateo}
  \state{CA}
  \country{USA}
}
\email{wyue0125@alumni.stanford.edu}

\renewcommand{\shortauthors}{Trovato et al.}

\begin{abstract}
In recent years, multi-modal multi-domain fake news detection has garnered increasing attention.  Nevertheless, this direction presents two significant challenges: \textbf{(1) Failure to Capture Cross-Instance Narrative Consistency:} existing models usually evaluate each news in isolation, fail to capture cross-instance narrative consistency, and thus struggle to address the spread of cluster-based fake news driven by social media; \textbf{(2) Lack of Domain-Specific Knowledge for Reasoning:} conventional models, which rely solely on knowledge encoded in their parameters during training, struggle to generalize to new or data-scarce domains (e.g., emerging events or niche topics). To tackle these challenges, we introduce \underline{\textbf{R}}etrieval-\underline{\textbf{A}}ugmented \underline{\textbf{M}}ultimodal \underline{\textbf{M}}odel for Fake News Detection (RAMM). 
First, RAMM employs a Multimodal Large Language Model (MLLM) as its backbone to capture cross-modal semantic information from news samples. Second, RAMM incorporates an Abstract Narrative Alignment Module. This component adaptively extracts abstract narrative consistency from diverse instances across distinct domains, aggregates relevant knowledge, and thereby enables the modeling of high-level narrative information.
Finally, RAMM introduces a Semantic Representation Alignment Module, which aligns the model’s decision-making paradigm with that of humans—specifically, it shifts the model’s reasoning process from direct inference on multimodal features to an instance-based analogical reasoning process.  Extensive experimental results on three public datasets validate the efficacy of our proposed approach. Our code is available at the following link: \href{https://github.com/li-yiheng/RAMM}{https://github.com/li-yiheng/RAMM}
\end{abstract}



\maketitle

\section{Introduction}
The popularity of social media platforms has fundamentally reshaped the landscape of information dissemination, fostering an environment where fake news can spread at an unprecedented speed and scale \cite{gonzalez2020social,keles2020systematic, lu2026dealt}. 
A particularly insidious form of fake news exploits the synergy between textual and visual content—by strategically employing engaging images to amplify false narratives \cite{pennycook2021psychology, kwon2013prominent,tong2025dapt}, it renders fabricated stories more credible and emotionally resonant. This phenomenon has driven the emergence of the critical research field of multimodal fake news detection, which seeks to develop computational methods for distinguishing fact from fiction through the joint analysis of text and images \cite{cui2019defend,zhou2015real,zhu2022generalizing,lu2026blind,zeng2025understand}.

However, there exists a crucial yet frequently overlooked challenge in current research: the domain specificity of news. For instance, the characteristics and distribution of fake news in the political domain exhibit significant differences from those in domains such as health, science, or entertainment. A model trained to detect political disinformation may exhibit poor performance when confronted with fabricated health advice, primarily due to such domain disparities. To address this critical gap, multimodal cross-domain fake news detection has emerged as a vital and highly promising research frontier, which aims to enhance the robustness and generalization ability of detection models across different news domains. Existing models have demonstrated outstanding performance. \cite{silva2021embracing} proposed a new framework that is capable of preserving both domain-specific and cross-domain knowledge within news records. KG-MFEND \cite{chen2023multi} develops sentence trees to capture the specific semantics of news across different domains.   Transm3 \cite{li2024apk} is enhanced by the Adaptive Pooling Kernel Convolutional Neural Network. MMDFND \cite{tong2024mmdfnd} extracts domain-specific semantics through a progressive hierarchical extraction network.
\begin{figure}[t]
    \centering
    \includegraphics[width=1\linewidth]{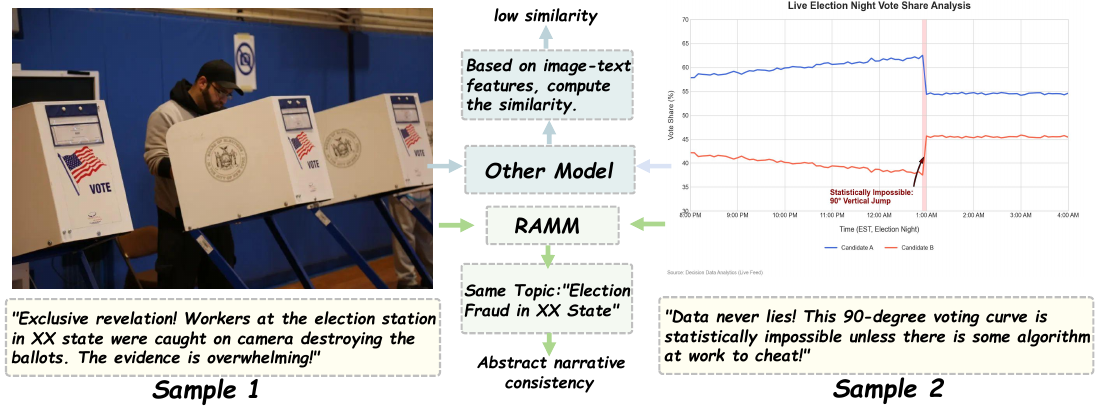}
    \caption{The issue of cluster-based propagation of fake news: different news items exhibit significant differences in their image-text content, yet all aim to persuade readers to believe the same claim-"Election Fraud".}
    \label{intro}
\end{figure}
Despite the aforementioned advancements, existing methods still face significant obstacles when deployed in real-world cross-domain scenarios, revealing critical limitations that hinder their effectiveness. We categorize these limitations into two major challenges:

\begin{itemize}
    \item \textbf{Failure to Capture Cross-Instance Narrative Consistency}: As shown in Figure \ref{intro}, fake news typically spreads in the form of "clusters"—multiple distinct news items (containing different images and texts) propagate the same underlying false narrative. Existing models process each news item independently and fail to perceive this higher-level consistency \cite{lu2025dammfnd}, thus struggling to identify a specific news item as part of a coordinated disinformation campaign (which serves as key evidence for ascertaining its falsity) \cite{subramanian2025overview, yi2025challenges}.

    \item \textbf{Lack of Domain-Specific Knowledge for Reasoning}: Detecting complex fake news requires not only evaluating text-image consistency but also reasoning based on domain-specific world knowledge \cite{raza2025fake, yu2025racmc}. For instance, verifying a claim about a new scientific discovery necessitates knowledge of established scientific principles \cite{mouratidis2025misinformation, ferdush2025cross, zeng2024mitigating}. Current models lack explicit mechanisms for acquiring and integrating such external domain-specific knowledge; instead, they often rely on spurious correlations within training data for learning, which results in poor generalization performance in unseen domains \cite{hussain2025fake}.

\end{itemize}

To address these fundamental challenges, we propose a \underline{\textbf{R}}etrieval-\underline{\textbf{A}}ugmented \underline{\textbf{M}}ultimodal \underline{\textbf{M}}odel for Fake News Detection (abbreviated as \textbf{RAMM}). This approach draws inspiration from the observation that human factcheckers do not evaluate news in isolation—instead, they instinctively retrieve and compare existing relevant information to identify overarching narratives and inconsistencies. By integrating a retrieval-augmentation mechanism into the detection pipeline based on Large Language Models (LLMs), RAMM translates this intuition into an actionable technical framework. Its core idea is to enrich the representation of the target news article by retrieving "homogeneous narrative sets" of semantically similar news items.
Specifically, this framework comprises three key components. RAMM employs a \textbf{M}ultimodal \textbf{L}arge \textbf{L}anguage \textbf{M}odel (MLLM) as its backbone to capture cross-modal semantic information from news samples. \textbf{A}bstract \textbf{N}arrative \textbf{A}lignment \textbf{M}odule (ANA) leverages LLMs to extract the core narrative from a given news item, acquires relevant news from in-domain and out-domain sources, synthesizes the information, and generates final, robust veracity predictions. \textbf{S}emantic \textbf{R}epresentation \textbf{A}lignment \textbf{M}odule (SRA), which aligns the model’s decision-making paradigm
with that of humans, it shifts the model’s reasoning process from direct inference on multimodal features to an instance-based analogical reasoning process.

Our contributions are as follows:

\begin{itemize}
    \item This study proposes \textbf{RAMM}, a novel retrieval-augmented framework for multimodal cross-domain fake news detection. As far as we are aware, this is the first research to explicitly equip  MLLMs with a retrieval mechanism to address the issues of narrative clusters and domain-specific knowledge gaps in fake news detection.

    \item It introduces a  \textbf{A}bstract \textbf{N}arrative \textbf{A}lignment \textbf{M}odule (ANA): leveraging the reasoning capabilities of LLMs, this module extracts high-level, abstract narrative information, freeing itself from reliance on surface-level features. And then identify and integrate information from news clusters with consistent underlying narratives, effectively bridging the semantic gap between raw content and veracity judgment.

    \item It presents a \textbf{S}emantic \textbf{R}epresentation \textbf{A}lignment \textbf{M}odule (SRA), redirecting the model’s reasoning process away from direct inference on multimodal features toward an instance-based analogical reasoning process like human.
    
    \item Extensive experiments are conducted on multiple publicly available datasets. The results demonstrate that the proposed RAMM framework outperforms state-of-the-art methods significantly, with outstanding performance particularly in challenging cross-domain scenarios—fully verifying its superior effectiveness and generalization ability.

\end{itemize}

\section{RELATED WORKS}
\subsection{LLMs for Fake News Detection}
The rapid advancement of Large Language Models (LLMs)\cite{wei2025igniting,wei2026beyond,liu2025coherency} and Multi-Agent System \cite{li2026decoding} has introduced a paradigm shift in the field of fake news detection. 
Recent approaches have explored more sophisticated ways to utilize the capabilities of LLMs\cite{zeng2026manipulation}. For example, FactAgent is an agentic approach that mimics the workflow of human fact-checkers by using an LLM to integrate its internal knowledge with information gathered from external tools to verify news claims \cite{li2024large}. This method allows for a more dynamic and comprehensive verification process that does not rely solely on the content of the article itself. Similarly, multi-agent LLM frameworks have been proposed to enhance detection by assigning specialized agents to different aspects of the verification task, such as analyzing source credibility, content style, and factual claims \cite{jeptoo2024enhancing}.
To enhance the interpretability and reasoning process of detection. Some frameworks use LLMs to produce justifications for their classification by reasoning through the evidence found \cite{papageorgiou2024survey}. A defense-based explainable framework was proposed where an LLM is prompted to generate justifications for both "real" and "fake" verdicts, and a final inference is made by modeling the defense between these competing arguments \cite{wang2024explainable}. Furthermore, some studies have shown that LLM-generated explanations can help human users in identifying fake news. By modeling the relationships between news, entities, and topics, their model can mine deeper semantic patterns indicative of fake news \cite{ma2024fake}. Another network for feature extraction with the advanced reasoning capabilities of LLMs to provide auxiliary explanations, thereby improving detection accuracy \cite{papageorgiou2025harnessing}.

\subsection{Multimodal Fake News Detection}

Owing to the remarkable achievements of deep learning and multimodal in diverse application domains \citep{li2025multi,lu2025dmmd4sr,li2023ultrare,zeng2026manipulation,cui2025diffusion}, an increasing number of studies have integrated deep learning models into multimodal fake news task\cite{zeng2024mitigating}.

Multimodal fake news detection is a method that utilizes various types of data, such as text, images, videos, audio, etc., to identify and detect fake news \cite{lakzaei2025neighborhood}. MVAE \cite{khattar2019mvae} proposes a novel multimodal variational autoencoder model capable of simultaneously processing multiple data sources. MCNN \cite{xue2021detecting} has designed a comprehensive framework that integrates feature extraction. It employs a consistency scoring mechanism to measure the degree of matching between features. \cite{chen2022cross}  designs a novel framework that enhances detection accuracy through learning from cross-modal ambiguity. MMDFND \cite{tong2024mmdfnd} extracts domain-specific semantics  and utilizes a pivot transformer network to integrate information from different modalities. \cite{jiang2025cross} enhanced few-shot  by transforming n-shot classification into  (n$\times$z)-shot problem. \cite{gong2025unseen} employs a reweighting strategy based on
modal confidence and propagation structure. \cite{yan2025mtpareto} designed hierarchical fusion network, defined three fusion levels to optimize multimodal fusion. The aforementioned methods overlook the complementarity and differences between different modalities during information integration. The simple fusion of modal features can lead to negative transfer on the model's performance.

\section{PROBLEM STATEMENT}We denote the sets of news by $\mathcal{U}$. Each sample $u \in \mathcal{U}$ of multimodal news is represented as \(\mathcal{N}_u = [I_u, T_u] \), where $I$ and $T$ respectively represent the image and text. The news in the dataset is categorized into $k$ classes, each assigned a domain label $ d_u \in \{ Domain_1, \ldots, Domain_k \} $. 
Given a news piece $u$ that incorporates both text and image information, and a domain label $d_u$, the objective of multi-modal multi-domain fake news detection is to determine the authenticity of the news piece.
\section{METHODOLOGY}
\begin{figure*}
    \centering
    \includegraphics[width=\linewidth]{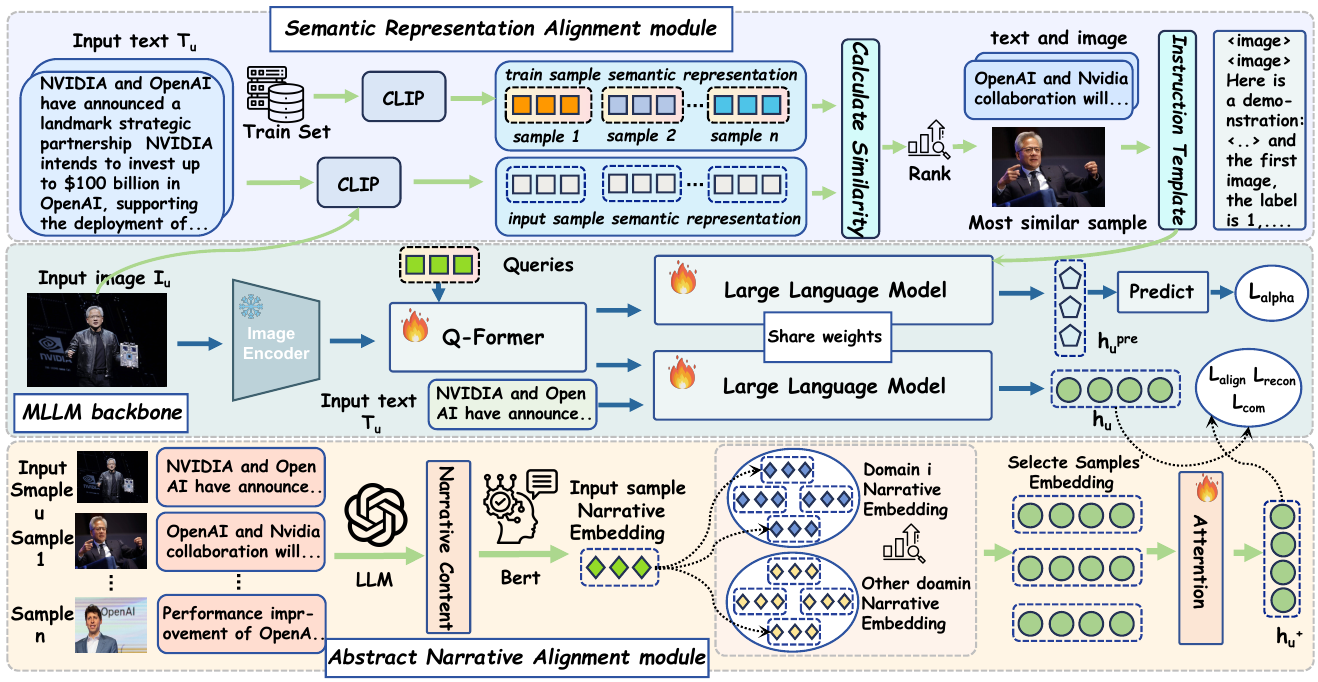}
    \caption{The architecture of RAMM. Semantic Representation Alignment Module selects examples based on image-text similarity, together with the target samples, are input into the MLLM backbone to extract features for discrimination. Abstract Narrative Alignment Module updates parameters in accordance with CIBL loss to capture high-level narrative correlations.}
    \label{fig:RAMM Framework}
\end{figure*}
In this section, we present a detailed description of the RAMM framework, as illustrated in Figure \ref{fig:RAMM Framework}. The RAMM framework is built upon a MLLM backbone and incorporates two primary modules: a Semantic Representation Alignment Module through multimodal similarity, and an Abstract Narrative Alignment Module refined via large language models. 

\subsection{MLLM Backbone}
In order to effectively identify disinformation in multimodal news, we have adopted the MLLM as the backbone, which consists of an image encoder, an LLM, and a Query Transformer (Q-Former). 
\subsubsection{Visual Feature Extraction}

We employ a pre-trained Vision Transformer (ViT), adopted from the BLIP model, as our visual backbone. An input image $I_u \in \mathbb{R}^{H \times W \times C}$ is initially partitioned into a sequence of fixed-size patches. These patches are then linearly projected into embeddings and combined with positional information. The resulting sequence is fed into the ViT encoder to produce a set of patch features.

To efficiently bridge the visual encoder with the language model, we introduce a Q-Former module. The Q-Former utilizes a small set of learnable queries to distill the most salient visual information from the ViT's output features via cross-attention.

Finally, a Multilayer Perceptron (MLP) projection layer maps the output from the Q-Former to the same dimension $d$ as the language model's text features. This yields the final image feature representation $V_u = \{v_1, v_2, \dots, v_m\}$, where each $v_i \in \mathbb{R}^{d}$. The entire process can be summarized as:
\begin{equation}\label{eq:visual_extraction}
    V_u = \text{MLP}(\text{Q-Former}(\text{ViT}(I_u)))
\end{equation}

\subsubsection{Textual Feature Extraction}
For the input text $T_u$, we use the tokenizer associated with  LLM to convert it into a sequence of tokens $\{t_1, t_2, \dots, t_n\}$. Then, we utilize the LLM's own word embedding layer to map this token sequence into a set of high-dimensional text feature vectors $A_u = \{a_1, a_2, \dots, a_n\}$, where $a_i \in \mathbb{R}^{d}$.

\subsubsection{Multimodal Feature Extraction}
We prepend the aligned visual feature $V_u$ as a "visual prompt" to the text feature sequence $A_u$, forming a unified multimodal input sequence:
\begin{equation}\label{fuse}
    E_u = \text{concat}(V_u, A_u) = \{v_1, \dots, v_m, a_1, \dots, a_n\}
\end{equation}
This fused sequence $E_u$ is directly fed into the LLM's Transformer architecture. 
To obtain a global representation for the entire image-text pair, we adopt the final hidden state of the last token as the aggregated feature vector $h_u$.
\begin{equation}\label{enc}
    h_u = \text{LLM}(E_u)_{[-1]}
\end{equation}
where $(\cdot)_{[-1]}$ denotes taking the last vector from the output sequence. 

\subsection{Abstract Narrative Alignment Module}
A primary challenge in fake news detection is its dissemination in event-specific "clusters", as shown in Figure \ref{intro}. Consequently, we propose the \textbf{A}bstract \textbf{N}arrative \textbf{A}lignment Module \textbf{(ANA)} to explicitly isolate and model the unifying deceptive narrative that connects seemingly disparate news items targeting the same event.

\subsubsection{Abstract narrative with LLMs}
\label{sec:narrative extrct}
Multimodal news data is frequently characterized by a composite of attributes, including core claims, implicit stances, and argumentative frames. These attributes encapsulate a wealth of abstract narrative semantics, which is critical for deepening a model's understanding of the news's persuasive intent. Given sample $u$'s information $\mathcal{N}_u$, we extract the abstract narrative for image and text.
These multi-modal information are then organized into a prompt format (denoted as ${\mathcal{P}}_{u}$). Next, we utilize a public LLM API to generate a textual summary $\mathcal{A}_{u}$ of the news' abstract narrative. This process can be formulated as follows:
\begin{equation}\mathcal{A}_{u}=\mathrm{LLM}(\mathcal{P}_{u})\end{equation}
Our goal is to perform contrastive learning based on news similarity. However, directly measuring similarity using the textual summary is impractical. Therefore, we employ a pretrained text embedding model $\mathcal{M}$ to extract and convert the semantic information contained in the textual narrative into embeddings, formatted as follows:
\begin{equation}\tilde{\mathbf{h}}_u=\mathcal{M}(\mathcal{A}_u)\end{equation}
where $\tilde{\mathbf{h}}_u\in\mathbb{R}^{\tilde{d}}$ represents the narrative embedding corresponding
to $\mathcal{N}_u$ and $\tilde{d}$ is the embedding size of the text embedding model $\mathcal{M}$. Specifically, $\mathcal{M}$ indicates SimCSE-RoBERTa \cite{gao2021simcse} and BGE \cite{xiao2024c} in this paper.
\subsubsection{Domain-Aware Homogeneous  Retrieval}
\label{sec:domain-aware retrive}
Once the narrative embeddings of news are obtained, similar news can be retrieved based on semantic similarity. Specifically, for a given news  $\mathcal{N}_u$, we calculate the cosine similarity between its narrative embedding $\tilde{\mathbf{h}}_u$ and the narrative embeddings of other news. News are then ranked in descending order according to the computed semantic similarity. To account for the domain-specificity of samples, we construct homogeneous sets of high-level narrative information. These sets are assembled by sourcing related news articles from both in-domain and out-of-domain sets. The top $k$ news are retrieved to construct the set of homogeneous news for sample $u$, denoted as $\overline{\mathcal{Q}}_u$ and $\tilde{\mathcal{Q}}_u$ respectively. This process is formulated as:
\begin{equation}\overline{\mathcal{Q}}_u=\{u^{\prime}\in\mathcal{U}\setminus\{u\}\mid\mathrm{rank}(\mathrm{cosine}_\text{sim}(\tilde{\mathbf{h}}_{u},\tilde{\mathbf{h}}_{u^{\prime}}))\leq k_{in},d_{u^{\prime}}=d_u\}\end{equation}
\begin{equation}\tilde{\mathcal{Q}}_{u}=\{u^{\prime}\in\mathcal{U}\setminus\{u\}\mid\mathrm{rank}(\mathrm{cosine}_\text{sim}(\tilde{\mathbf{h}}_{u},\tilde{\mathbf{h}}_{u^{\prime}}))\leq k_{out},d_{u^{\prime}}\neq d_u\}\end{equation}
where $u^{\prime}\in\mathcal{U}\setminus\{u\}$ denotes the set of all news except $u$, $k_{in}$ and $k_{out}$ denote the number of in-domain and out-of-domain retrieved similar news, respectively.

Subsequently, we combine the in-domain and out-of-domain sets, yielding the final homogeneous news collection  $\mathcal{Q}_u = \overline{\mathcal{Q}}_u \cup \tilde{\mathcal{Q}}_u$.

\subsubsection{Narrative-Enhanced Fusion}
After identifying homogeneous news for each news item based on narrative information, a straightforward approach for contrastive learning is to use a homogeneous news item \( u' \in \mathcal{Q}_u \) as the positive sample for news \( u \). In this scenario, the outputs of the Large Language Model (LLM), \( \mathbf{h}_{u'} \) and \( \mathbf{h}_u \), will be directly brought closer together. However, semantic similarity is computed in the text embedding space, which may differ from the multimodal space of the original samples. This discrepancy may lead to representation misalignment. 

To address this issue, we propose using retrieved news as a candidate set and designing a learnable method to synthesize an appropriate positive sample for contrastive learning from this candidate set. Specifically, we first map the semantic representations of news through a learnable adapter. Then, in the mapped space, we employ an attention mechanism, where the current news serves as a query to compute the probability \( p_{u,u'} \) that each candidate news \( u' \in \mathcal{Q}_u \) is suitable as the positive sample for the current news \( u \). This process is formulated as:
\begin{equation}\begin{aligned}
 & w_{u,u^{\prime}}=\mathrm{LeakyReLU}(\mathbf{a}^{\top}[\mathbf{W}\tilde{\mathbf{h}}_{u}\|\mathbf{W}\tilde{\mathbf{h}}_{u^{\prime}}]), \\
 & p_{u,u^{\prime}}=\mathrm{softmax}_{u^{\prime}}(w_{u,u^{\prime}})=\frac{\exp(w_{u,u^{\prime}})}{\sum_{u_{k}\in\mathcal{N}_{u}}\exp(w_{u,u_{k}})},
\end{aligned}\end{equation}
where $\mathbf{W}\in\mathbb{R}^{d\times\tilde{d}}$ is a learnable weight matrix that adapts the
semantic embeddings to the representation space of the MLLM model, and $\|$ denotes the concatenation operation. $\mathbf{a}\in\mathbb{R}^{2d}$ represents a single-layer neural network used to generate the attention score, with the LeakyReLU activation function. The softmax function is employed to transform
the coefficients into probabilities. Based on this, we perform a
weighted sum of $\mathbf{h}_{u'}$ for all $u' \in \mathcal{Q}_u$ to obtain the composite positive
contrastive sample $\mathbf{h}_{u}^+$ for $\mathbf{h}_{u}$:
\begin{equation}\mathbf{h}_{u}^{+}=\sum_{u^{\prime}\in\mathcal{N}_{u}}p_{u,u^{\prime}}\mathbf{h}_{u^{\prime}},\end{equation}
where $\mathbf{h}_{u'} \in \mathbb{R}^{d} $ is the MLLM’s output multinmodal representation for $u'$,as defined in Equation~\eqref{enc}.
\subsubsection{Common Information Bottleneck Loss (CIBL)}
\label{sec:CIBL}
Most existing feature optimization methods rely on single contrastive learning to align features. However, if one of the features is noisy or becomes blurred due to averaging, such forced alignment will impair the valuable personalized information in $\mathbf{h}_{u}$. Accordingly, we propose a new optimization method based on \textbf{C}ommon \textbf{I}nformation theory and \textbf{I}nformation \textbf{B}ottleneck theory (abbreviated as \textbf{CIBL}).

The feature $\mathbf{h}_{u}$ of news $u$ and its synthetic positive sample $\mathbf{h}_{u}^+$ are both noisy observations of a certain latent and more pristine "common abstract narrative" $\mathcal{Z}_{u}$. Therefore, we opt to jointly distill this latent common abstract narrative $\mathcal{Z}_{u}$ from $\mathbf{h}_{u}$ and $\mathbf{h}_{u}^+$.
Specifically, we feed $\mathbf{h}_u$ and $\mathbf{h}_{u}^+$ into a joint variational encoder $q_{\phi}$, which parameterizes a Gaussian posterior distribution $q_{\phi}(\mathbf{z}_u | \mathbf{h}_u, \mathbf{h}_u^+)$. This encoder consists of two $MLP$ networks $f_{\mu}$ and $f_{\sigma}$, which respectively predict the mean $\boldsymbol{\mu}_u$ and logarithmic variance $\log \boldsymbol{\sigma}_u^2$:
\begin{equation}
    \boldsymbol{\mu}_u, \log \boldsymbol{\sigma}_u^2 = f_{\mu}([\mathbf{h}_u || \mathbf{h}_{u}^+]), f_{\sigma}([\mathbf{h}_u || \mathbf{h}_{u}^+])
\end{equation}

Then, we obtain the latent common narrative representation $\mathcal{Z}_u$ by sampling from this distribution via the reparameterization trick:
\begin{equation}
    \mathcal{Z}_u = \boldsymbol{\mu}_u + \boldsymbol{\sigma}_u \odot \boldsymbol{\epsilon}, \quad \text{where} \ \boldsymbol{\epsilon} \sim \mathcal{N}(0, \mathbf{I})
\end{equation}
where $\odot$ denotes element-wise multiplication.

To ensure that the latent representation $\mathcal{Z}_u$ accurately captures the common abstract narrative information between $\mathbf{h}_{u}$ and $\mathbf{h}_{u}^+$, while maintaining its conciseness and robustness, we have designed three complementary self-supervised loss functions.

Firstly, within the framework of common information theory, $\mathcal{Z}_u$ must contain sufficient information to account for $\mathbf{h}_{u}^+$. Therefore, we compute the alignment loss $\mathcal{L}_{\text{align}}$ by treating ($\mathcal{Z}_u$, $\mathbf{h}_{u}^+$) as a positive sample pair.
\begin{equation}
    \mathcal{L}_{\text{align}} = -\log \frac{\exp(\text{sim}(\mathcal{Z}_u, \mathbf{h}_{u}^+) / \tau)}{\exp(\text{sim}(\mathcal{Z}_u, \mathbf{h}_{u}^+) / \tau) + \sum_{j \neq u} \exp(\text{sim}(\mathcal{Z}_u, \mathbf{h}_j^+) / \tau)}
\end{equation}
where $\text{sim}(\cdot, \cdot)$ denotes cosine similarity, and $\tau$ is the temperature hyperparameter. This ensures that $\mathcal{Z}_u$ contains valuable signals related to common group narrative, rather than some irrelevant noise.

Secondly, To avoid the loss of the abstract narrative information of news $u$ in $\mathcal{Z}_u$ caused by mere alignment with $\mathbf{h}_{u}^+$, we introduce a decoder $g_{\psi}$ and enforce it to reconstruct the original $\mathbf{h}_{u}$ from $\mathcal{Z}_u$, thereby ensuring that $\mathcal{Z}_u$ contains key information about $\mathbf{h}_{u}$:
\begin{equation}
    \mathcal{L}_{\text{recon}} = \left\| g_{\psi}(\mathbf{z}_u) - \mathbf{h}_u \right\|_2^2
\end{equation}

Lastly, to enable the model to discard non-essential, redundant, or noisy information in $\mathbf{h}_{u}$ and $\mathbf{h}_{u}^+$, while retaining only the most core and essential shared narrative signals in $\mathcal{Z}_u$, we accomplish this by means of KL divergence.
\begin{equation}
    \mathcal{L}_{\text{compress}} = \text{KL}(q_{\phi}(\mathcal{Z}_u | \mathbf{h}_u, \mathbf{h}_u^+) \| p(\mathbf{z})) = \frac{1}{2} \sum_{i=1}^{d_z} (\sigma_{u,i}^2 + \mu_{u,i}^2 - 1 - \log \sigma_{u,i}^2)
\end{equation}
Furthermore, we provide detailed theoretical proofs to demonstrate that our scheme can effectively alleviate the problem of representation collapse, which can be found in Appendix \ref{sec:theory}.
\subsection{Semantic Representation Alignment Module}
In the task of vetting misinformation, humans tend to leverage concrete prior experiences. 
Our proposed \textbf{S}emantic \textbf{R}epresentation \textbf{A}lignment \textbf{M}odule \textbf{(SRA)} serves as a functional analogue to this cognitive mechanism. Fundamentally, it pivots the model's decision-making paradigm from direct inference on multimodal features to a process of instance-based analogical reasoning.

As shown in Figure 2, for the given image-text pairs $\mathcal{N} = [I, T]$, we first obtain their corresponding
embeddings by CLIP: 
\begin{equation}
f_{\text{CLIP-I}} = \operatorname{CLIP}_{\text{vis}}(I), \quad f_{\text{CLIP-T}} = \operatorname{CLIP}_{\text{text}}(T)
\end{equation}
where $\operatorname{CLIP}_{\operatorname{vis}}$ and $\operatorname{CLIP}_{\operatorname{text}}$ are the visual and textual encoder of CLIP.

For each sample $\langle f_{\text{CLIP-I}}, f_{\text{CLIP-T}}\rangle$,we calculate the cosine similarity of image and text modalities separately with the samples in the training set $D_{train}$:
\begin{equation}Sim_{v}(i)=\frac{f_{\text{CLIP-I}}\cdot\mathbb{V}}{|f_{\text{CLIP-I}}||\mathbb{V}|},Sim_{t}(i)=\frac{f_{\text{CLIP-T}}\cdot\mathbb{T}}{|f_{\text{CLIP-T}}||\mathbb{T}|}\end{equation}
where $\mathbb{V}$ and $\mathbb{T}$ are the image and text embeddings from the training set $D_{train}$.

Finally, we select the sample with the highest average similarity score as the corresponding demonstration:
\begin{equation}Demon(i)=\arg\max\frac{Sim_v(i)+Sim_t(i)}{2}\end{equation}
where $Demon(i)$ is the similarity score of demonstration of i-th image-text sample. 

Then the retrieved demonstration and the sample, after being processed through the instruction template, are input into LLM together, as shown in Figure \ref{fig:RAMM Framework}. In contrast to the feature concatenation approach of Equation~\eqref{fuse}, our method integrates visual information by embedding dedicated \texttt{<image>} tokens directly into the instruction template. This results in an input sequence $\{t_1,t_2,\dots,   t_{\text{image}},\dots, \ t_{\text{image}}, \dots, t_n\}$ and a corresponding feature matrix $E_T = \{e_1, e_2, \dots, e_{\text{image}}, \dots, e_{image} \dots, e_n\}$. The placeholder image embeddings, $e_{\text{image}}$, are then substituted with the features of the retrieved demonstration and the input sample (extracted via Equation~\eqref{fuse}), respectively, to form the final fused matrix, $E_u^{fuse}$. 
\begin{equation}
    \mathbf{h}_u^{pre} = \text{LLM}(E_u^{fuse})_{[-1]}
\end{equation}
The final representation $\mathbf{h}_u^{pre}$ is then passed through a final classification layer with a sigmoid activation function to produce the probability of the news being fake, $\hat{y}_u = \sigma(\text{Linear}(\mathbf{h}_u^{pre}))$.
The classification loss $\mathcal{L}_{alpha}$ is the Binary Cross-Entropy (BCE) loss:
\begin{equation}
    \mathcal{L}_{alpha} = - \frac{1}{N} \sum_{i=1}^{N} \left[ y_i \log(\hat{y}_i) + (1-y_i) \log(1-\hat{y}_i) \right].
\end{equation}
\paragraph{Note} To prevent information leakage, the SRA and ANA modules retrieve relevant samples from the training set solely based on the query during both the training and inference phases.
\subsection{Optimization}
The overall model is trained by minimizing a composite loss function $\mathcal{L}_{total}$, which is shown as below:
\begin{equation}
    \mathcal{L}_{total} = \mathcal{L}_{alpha} + \lambda_1 \mathcal{L}_{align} + \lambda_2 \mathcal{L}_{recon} + \lambda_3 \mathcal{L}_{compress}
\end{equation}
where $\lambda_{1,2,3}$ are hyperparameter to CIBL  loss (\ref{sec:CIBL}) respectively.

\section{EXPERIMENTS}
In this section, we conducted an extensive series of experiments to rigorously evaluate the performance of RAMM.  Our aim was to address the following research questions:

\begin{itemize}
    \item  \textbf{(Effectiveness) RQ1:} Does our proposed RAMM outperform state-of-the-art fake news detection baselines?
    \item  \textbf{(Ablation) RQ2:} How do the key components within RAMM contribute to its overall performance?
    \item  \textbf{(Sensitivity) RQ3:} How sensitive is RAMM to its key hyperparameters?
    \item  \textbf{(Robustness) RQ4:} Can RAMM maintain robust  in zero-shot strategy?
    \item  \textbf{(Efficiency) RQ5:} Can RAMM maintain robustness when faced with noise interference?
    \item  \textbf{(Visualization) RQ6:} Can RAMM learn discriminative representations for fake news detection?
\end{itemize}

\subsection{Experimental Settings}

\subsubsection{Datasets}
Following previous work \cite{tong2024mmdfnd}, our model is evaluated on three real-world datasets: two Chinese multi-domain datasets, Weibo \cite{wang2018eann} and Weibo-21 \cite{nan2021mdfend}, and one English multi-domain dataset, FineFake \cite{zhou2024finefake}. The Weibo and Weibo-21 datasets are divided into nine domains, while FineFake contains six domains and one uncategorized domain. We use the same data partitioning method as MMDFND \cite{tong2024mmdfnd} and KEAN \cite{zhou2024finefake}.

\subsubsection{Baselines}
To comprehensively assess the performance of our proposed model, we compare it against baseline methods organized into three categories:  (1) single-modal multi-domain methods, including:\textbf{MDFEND} \cite{nan2021mdfend}, \textbf{M$^3$DFEND} \cite{zhu2022memory} and \textbf{PLDFEND} \cite{peng2023prompt}; (2) multi-modal multi-domain methods, including: \textbf{KATMF} \cite{song2021knowledge}, \textbf{MMDFND} \cite{tong2024mmdfnd}, \textbf{DAMMFND} \cite{lu2025dammfnd}, \textbf{IMOL} \cite{zeng2025imol} and \textbf{MMHT} \cite{yang2025macro}; (3) MLLM methods, including: \textbf{GSFND} \cite{tong2025generate}, \textbf{SNIFFER} \cite{qi2024sniffer}, \textbf{NRFE} \cite{zhang2025llms}, \textbf{BLIP2} \cite{li2023blip}.
\subsubsection{Implementation Details}
We implement our model using PyTorch 2.0 and conduct all experiments on NVIDIA A100  GPUs. We employ BLIP-2 with ViT-L/14 as the visual encoder and Qwen2.5-7B and Llama3-8B as the LLM backbone for Chinese and English datasets respectively. Qwen3-MAX is used to generate abstract narrative. The Q-Former uses 32 learnable queries with hidden dimension d=768. The text embedding model M is SimCSE-RoBERTa-large with embedding dimension $\tilde{d}=1024$. We train the model for 3 epochs using AdamW optimizer with learning rate 1e-4, weight decay 0.01, and cosine learning rate scheduler with 100 warmup steps. The batch size is set to 4. We retrieve k=5 similar samples ($k_{in} = 3$ and $k_{out}=2$) for narrative alignment. The balancing hyperparameter $\lambda_1$, $\lambda_2$ and $\lambda_3$ for the CIBL loss is set to 0.2, 0.1 and 0.2 respectively. 


\begin{table*}[!ht]
\small 
\centering
\renewcommand{\arraystretch}{1.0} 
\caption{Comparison between RAMM and the latest multi-domain fake news detection methods on Weibo, Weibo-21 and FineFake. *: open-source. The best results are in \textbf{bold}, and the second-best results are \underline{underlined}. The $t$-tests validate the significance of performance improvements with $p$-value $\leq 0.05$.}
\begin{tabular}{c p{1.5cm}<{\centering} p{0.7cm}<{\centering} p{0.8cm}<{\centering} p{0.8cm}<{\centering} p{0.7cm}<{\centering} p{0.7cm}<{\centering} p{0.7cm}<{\centering} p{0.7cm}<{\centering} p{0.7cm}<{\centering} p{0.8cm}<{\centering} p{0.7cm}<{\centering} p{0.7cm}<{\centering} p{0.7cm}<{\centering}}
\hline 
 \rowcolor{tableheaderblue!30} \multirow{2}{*}{}   &  & Sci. & Mil.  & Edu.   & Soc. & Pol. & Hlth. & Fin. & Ent. & Dis.  & \multicolumn{3}{c}{Overall} \\  \cline{12-14} 

      \rowcolor{tableheaderblue!30}                    &       \textbf{M}ethod                  &       &   Con.     &  Unc.      &   Soc.    &  Pol.     &   Hlth.     &  Fin.     &  Ent.     &  Int.      & F1    & Acc   & AUC   \\ \hline
\multirow{7}{*}{\rotatebox{90}{Weibo}}    
                          & MDFEND*                  & 0.772 & 0.914  & 0.895  & 0.905 & 0.761 & 0.876  & 0.810 & 0.879 & 0.876  & 0.903 & 0.903 & 0.966 \\
                          & M$^3$DFEND*              & 0.794 & 0.901  & 0.925  & 0.910 & 0.768 & 0.861  & 0.901 & 0.897 & 0.878  & 0.927 & 0.927 & 0.970 \\
                          & PLDFEND                  & 0.840 & 0.915  & 0.882  & 0.927 & \underline{0.921} & 0.920  & 0.926 & 0.898 & 0.447  & 0.926 & 0.929 & 0.975 \\ 
                          & KATMF                    & 0.833 & 0.906  & 0.921  & 0.897 & 0.825 & 0.895  & 0.901 & 0.906 & 0.892  & 0.931 & 0.931 & 0.968 \\
                          & MMDFND*                  & 0.826 & 0.909  & 0.939  & 0.941 & 0.732 & 0.915  & 0.914 & 0.919 & 0.886  & 0.933 & 0.933 & 0.973 \\
                          & DAMMFND*           & 0.853 & 0.911 & \textbf{0.956} & \underline{0.943} & 0.822 & \underline{0.939} & \textbf{0.937} & \textbf{0.956} & \underline{0.928} & \underline{0.943} & \underline{0.944} & \underline{0.983} \\
                          & IMOL*                  & \underline{0.916} & \textbf{0.929}  & 0.937  & \textbf{0.949} & 0.812 & 0.915  & 0.924 & 0.935 & 0.921  & 0.941 & 0.942 & 0.979 \\
                          & MMHT                  & 0.896 & 0.913  & 0.941  & 0.935 & 0.879 & 0.920  & 0.915 & 0.952 & 0.902  & 0.939 & 0.941 & 0.979 \\
                          \cline{2-14} 

\rowcolor{yellow!20}        & \textbf{RAMM} & \textbf{0.935} & \underline{0.917} & \underline{0.952} & 0.937 & \textbf{0.942} & \textbf{0.966} & \underline{0.935} & \underline{0.954} & \textbf{0.931} & \textbf{0.959} & \textbf{0.961} & \textbf{0.989} \\\hline
\multirow{9}{*}{\rotatebox{90}{Weibo-21}} 
                          & MDFEND*                  & 0.832 & 0.936  & 0.893  & 0.895 & 0.888 & 0.938  & 0.897 & 0.904 & 0.902  & 0.914 & 0.914 & 0.969 \\
                          & M$^3$DFEND*              & 0.831 & 0.948  & 0.901  & 0.906 & 0.885 & \underline{0.943}  & 0.903 & 0.929 & 0.891  & 0.920 & 0.920 & 0.976 \\     
                          & PLDFEND                 & 0.910 & \textbf{0.959} & 0.901  & 0.911 & 0.888 & 0.940  & 0.885 & 0.936 & 0.887  & 0.925 & -     & -     \\     
                          & KATMF                    & 0.912 & 0.930  & 0.911  & 0.898 & 0.899 & 0.917  & 0.873 & 0.935 & 0.900  & 0.925 & 0.929 & 0.974 \\     
                          & MMDFND*                  & 0.935 & 0.955  & 0.854  & 0.942 & 0.963 & 0.923  & 0.881 & \underline{0.961} & 0.917 & 0.940 & 0.940 & 0.976 \\
                          & DAMMFND*           & 0.932 & 0.953 & \underline{0.932} & 0.917 & \textbf{0.982} & 0.919 & 0.941 & \textbf{0.990} & \textbf{0.945} & \underline{0.947} & \underline{0.947} & \underline{0.985} \\
                          & IMOL*           & 0.857 & 0.921 & \textbf{0.954} & \underline{0.947} & 0.852 & 0.932 & \underline{0.945} & 0.953 & 0.925 & 0.943 & 0.944 & 0.981 \\
                          & MMHT                  & \underline{0.942} & 0.949  & 0.929  & 0.937 & 0.956 & 0.915  & 0.914 & 0.929 & 0.923  & 0.945 & 0.945 & 0.983 \\
                          \cline{2-14} 
\rowcolor{yellow!20}        & \textbf{RAMM} & \textbf{0.962} & \underline{0.956} & 0.927 & \textbf{0.961} & \underline{0.974} & \textbf{0.979} & \textbf{0.983} & 0.959 & \underline{0.937} & \textbf{0.968} & \textbf{0.969} & \textbf{0.991} \\ \hline
\multirow{9}{*}{\rotatebox{90}{FineFake}} 
                          & MDFEND*                  & -     & -      & -      & -     & -     & -      & -     & -     & -      & 0.781 & 0.788 & -     \\
                          & M$^3$DFEND*              & -     & -      & -      & -     & -     & -      & -     & -     & -      & 0.772 & 0.781 & -     \\     
                          & PLDFEND                  & -     & 0.724  & 0.638  & 0.785 & 0.766 & 0.777  & 0.829 & 0.835 & -      & 0.789 & 0.790 & 0.873 \\     
                          & KATMF                    & -     & 0.727  & 0.701  & 0.758 & 0.756 & \underline{0.805}  & 0.839 & 0.846 & -      & 0.782 & 0.786 & 0.869 \\     
                          & MMDFND*                   & -     & \underline{0.747} & 0.740  & 0.760 & 0.768 & 0.767  & 0.831 & 0.839 & -      & 0.786 & 0.789 & 0.868 \\
                          & DAMMFND*           & -     & 0.741 & \underline{0.746} & 0.787 & \underline{0.776} & 0.803 & 0.838 & \underline{0.853} & -      & 0.792 & 0.794 & 0.878 \\
                          & IMOL*                  & - & \textbf{0.750}  & 0.736  & \underline{0.792} & 0.774 & \textbf{0.809}  & \underline{0.842} & 0.851 & -  & \underline{0.793} & \underline{0.795} & \underline{0.880} \\
                          & MMHT                  & - & 0.739  & 0.735 & 0.775 & 0.770 & 0.795  & 0.837 & 0.845 & -  & 0.789 & 0.791 & 0.873 \\
                          \cline{2-14} 
\rowcolor{yellow!20}       & \textbf{RAMM}           & -     & 0.734 & \textbf{0.757} & \textbf{0.799} & \textbf{0.792} & 0.793 & \textbf{0.850} & \textbf{0.861} & -      & \textbf{0.809} & \textbf{0.811} & \textbf{0.892} \\ \hline
\end{tabular}

\label{multi-domian table}
\end{table*}

\begin{table}[t]
\centering
\caption{Comparison between RAMM and the latest MLLM methods on Weibo and FineFake datasets.}
\begin{tabular}{>{\centering\arraybackslash}p{0.5cm}l>{\centering\arraybackslash}p{0.5cm}l>{\centering\arraybackslash}p{1.2cm}ccc}
\hline
\rowcolor{tableheaderblue!30} \multicolumn{2}{c}{\textbf{Datasets}} & \multicolumn{2}{c}{\textbf{Method}} & \textbf{Accuracy} & \textbf{F1 score} & \textbf{AUC}\\\hline
\multicolumn{2}{c}{\multirow{5}{*}{Weibo}}     & \multicolumn{2}{c}{BLIP2*}                   & 0.872                     & 0.871          & 0.943          \\
\multicolumn{2}{c}{}                          & \multicolumn{2}{c}{SNIFFER*}                    & 0.903                     & 0.902          & 0.957          \\
\multicolumn{2}{c}{}                          & \multicolumn{2}{c}{GSFND}                    & 0.938                     & 0.937          & 0.975          \\
\multicolumn{2}{c}{}                          & \multicolumn{2}{c}{NRFE}                    & 0.945            & 0.944 & 0.978 \\ \cline{3-7} 
\rowcolor{yellow!20} \multicolumn{2}{c}{}                         & \multicolumn{2}{c}{\textbf{RAMM}}                    & \textbf{0.961}            & \textbf{0.959} & \textbf{0.989} \\ \hline
\multicolumn{2}{c}{\multirow{5}{*}{FineFake}}  & \multicolumn{2}{c}{BLIP2*}                   & 0.757                     & 0.756          & 0.838          \\
\multicolumn{2}{c}{}                          & \multicolumn{2}{c}{SNIFFER*}                   & 0.781                     & 0.770           & 0.852          \\
\multicolumn{2}{c}{}                          & \multicolumn{2}{c}{GSFND}                    & 0.785                     & 0.783          & 0.865          \\
\multicolumn{2}{c}{}                          & \multicolumn{2}{c}{NRFE}                    & 0.799            & 0.797 & 0.881 \\ \cline{3-7} 
\rowcolor{yellow!20} \multicolumn{2}{c}{}                          & \multicolumn{2}{c}{\textbf{RAMM}}                    & \textbf{0.811}            & \textbf{0.809} & \textbf{0.892}\\ \hline
\end{tabular}
\label{mllm table}
\end{table}

\subsection{Overall Performance (RQ1)}
\begin{table}[b]
\centering
\small
\renewcommand{\arraystretch}{1.0}
\setlength{\tabcolsep}{5pt}
\caption{Ablation study of RAMM on three datasets. We report the F1-score.}
\begin{tabular}{l c c c}
\toprule
\textbf{Variant} & \textbf{Weibo} & \textbf{Weibo-21} & \textbf{FineFake} \\
\midrule
\textbf{RAMM (Full Model)} & \textbf{0.959} & \textbf{0.968} & \textbf{0.809} \\
\midrule
\rowcolor{gray!0} 
\multicolumn{4}{l}{\textit{On Similar Sample Selection:}} \\
-SRA & 0.937 {\color{red}(-2.2\%)} & 0.947 {\color{red}(-2.1\%)} & 0.785 {\color{red}(-2.4\%)} \\
-ANA (In-domain. only) & 0.947 {\color{red}(-1.2\%)} & 0.955 {\color{red}(-1.3\%)} & 0.791 {\color{red}(-1.8\%)} \\
-ANA (CLIP. select) & 0.941 {\color{red}(-1.8\%)} & 0.952 {\color{red}(-1.6\%)} & 0.787 {\color{red}(-2.2\%)} \\
\midrule
\rowcolor{gray!0} 
\multicolumn{4}{l}{\textit{On Information Bottleneck and Optimization:}} \\
-CIBL (Force. Align)& 0.943 {\color{red}(-1.6\%)} & 0.951 {\color{red}(-1.7\%)} & 0.794 {\color{red}(-1.5\%)} \\
-CIBL (Simple. Loss) & 0.940 {\color{red}(-1.9\%)} & 0.953 {\color{red}(-1.5\%)} & 0.793 {\color{red}(-1.6\%)} \\
\bottomrule
\end{tabular}

\label{tab:ablation}
\end{table}
Table \ref{multi-domian table} and Table \ref{mllm table} present the overall performance comparison between RAMM and various baseline methods. The results clearly demonstrate the superiority of our proposed approach across all three benchmark datasets.

\textbf{(1) RAMM achieves new state-of-the-art performance.} As shown, RAMM consistently outperforms all baselines in terms of overall F1, Accuracy, and AUC. For instance, on Weibo-21, RAMM achieves an F1-score of 0.968, surpassing the previous best method, DAMMFND, by 2.1\%. This significant improvement is attributed to RAMM's ability to mitigate negative transfer by judiciously selecting and integrating knowledge from the most beneficial higher-order narratives, thereby generating more robust and generalized representations.

\textbf{(2) Multimodal multi-domain methods demonstrate superior performance.} The results also indicate that multimodal methods (e.g., MMDFND, DAMMFND, RAMM) significantly outperform single-modal methods (e.g., PLDFEND, M$^3$DFEND). This highlights the critical importance of leveraging cross-modal information for detecting fake news. RAMM further advanced this process by not only leveraging multimodal cues but also explicitly modeling the narrative connections between similar samples from different domains. This made knowledge transfer more effective and ultimately led to its outstanding performance.

\textbf{(3) MLLM methods demonstrate superior contextual understanding.} The fundamental difference between traditional multimodal multi-domain fake news detection models and MLLMs lies in their core logic for information processing and fake content identification. MLLMs have shifted towards deeper-level "semantic understanding"—a transition that endows them with significant advantages in accuracy, generalization ability, and robustness. Building on this foundation, RAMM leverages high-level narrative information extracted from similar samples to enhance the domain generalization and reasoning performance of MLLMs, enhancing the ability to capture subtle clues.

\subsection{Ablation Study (RQ2)}
\label{sec:ablation}
We conduct a series of ablation experiments to validate the effectiveness of each key component in RAMM. The results, summarized in Table \ref{tab:ablation}, demonstrate that every module contributes positively to the final performance.




\textbf{Effect of Similar Sample Selection.} We design three variants to analyze the core selection mechanism. (1) \textbf{-SRA} removes SRA module, which compels the model to shift from analogical reasoning to guidance based on multimodal features . (2) \textbf{-ANA (In-domain. only)} replaces our Domain-Aware Homogeneous Retrieval (\ref{sec:domain-aware retrive}) with the approach of only retrieving in-domain similar samples. (3) \textbf{-ANA (CLIP. select)} removes narrative information extraction (\ref{sec:narrative extrct}), homogeneous retrieval and information fusion are performed based on the image-text similarity of CLIP across different samples.
Across all datasets, the significant performance degradation of the ``-SRA'' variant confirms our hypothesis: the approach based on analogical reasoning significantly enhances the semantic understanding capabilities of MLLMs.  
Similarly, the ``-ANA (CLIP select)'' variant confirms the core issue of previous models: retrieving samples solely based on feature similarity fails to capture the hidden high-level narrative correlations behind clusters under the "cluster propagation" mode, resulting in a decline in reasoning performance. 
The lesser, yet still notable, decline in the ``-ANA (In-domain. only) '' variant validates the importance of modeling the out-domain relationships for generalization performance.

\textbf{Effect of Information Bottleneck and Optimization.} We analyze the CIBL  strategy (\ref{sec:CIBL}) with two variants. (1) \textbf{-CIBL (Force. Align)} remove variational encoder and replaces "general abstract description" $\mathcal{Z}_{u}$ with $\mathbf{h}_{u}$ to achieve forced alignment.  (2) \textbf{-CIBL (Simple. Loss)} removes $\mathcal{L}_{\text{compress}}$, $\mathcal{L}_{\text{align}}$ and $\mathcal{L}_{\text{recon}}$, only use  contrastive loss for optimization. The performance degradation in both cases underscores the benefits of our designs. The results for ``-CIBL (Force. Align)'' show that forced alignment will impair the valuable personalized information in  $\mathbf{h}_{u}$. Similarly, the results for ``-CIBL (Simple. Loss)'' demonstrate that multiple loss will capture robust representation and discard non-essential or noisy information.

\subsection{Parameter Sensitivity Analysis (RQ3)}
\label{sec:param_sensitivity}
We analyzed the sensitivity of RAMM to four key hyperparameters on the Weibo dataset:  the number of selected in-domain samples $k_{in}$, the number of selected in-domain samples $k_{out}$, the loss balancing coefficient $\lambda_1$ and $\lambda_2$.

\begin{figure}[t]
  \centering
  \includegraphics[width=0.9\linewidth]{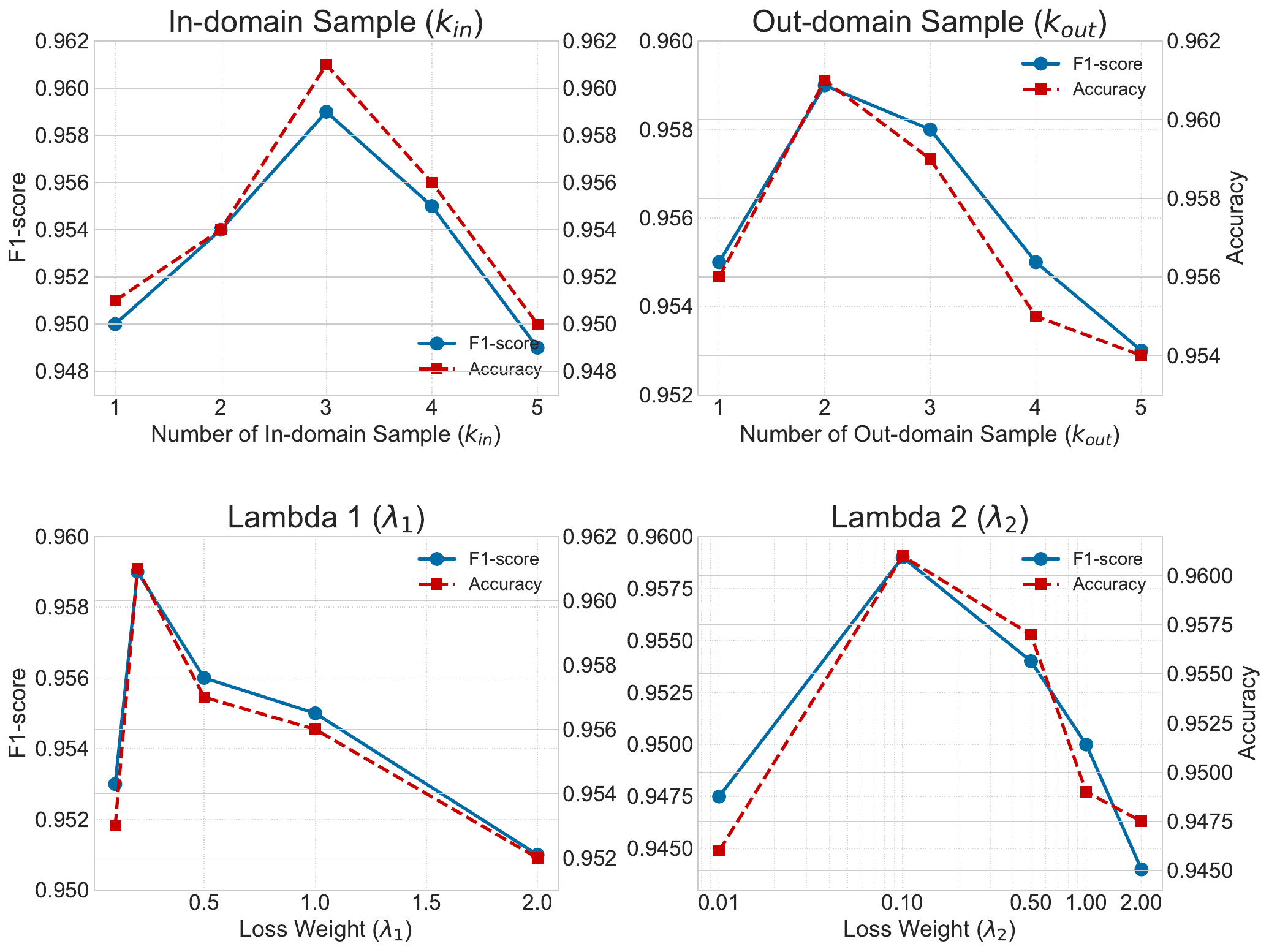}
  \caption{Parameter sensitivity analysis of four key hyperparameters on the Weibo dataset, evaluated by F1-score and Accuracy.}
  \label{fig:param_sensitivity}
  \vspace{-6pt}
\end{figure}

 As shown in Figure \ref{fig:param_sensitivity}, RAMM demonstrates robust performance across a reasonable range for all tested hyperparameters. For instance, performance peaks when the number of selected in-domain samples $k_{in}$ is 3 and the number of selected out-domain samples $k_{out}$ is 2, indicating that a moderate number of retrived samples can effectively capture domain characteristics and that fusing knowledge from a few highly relevant domains is more effective than including too many. The model is relatively stable with loss balancing coefficient from 0.1 to 0.3. This overall robustness reduces the burden of extensive hyperparameter tuning and demonstrates the stability of our proposed framework.
\subsection{Domain Generalization Analysis (RQ4)}
\label{sec:neg_transfer}
\begin{figure}[b]
  \centering
  \includegraphics[width=0.9\linewidth]{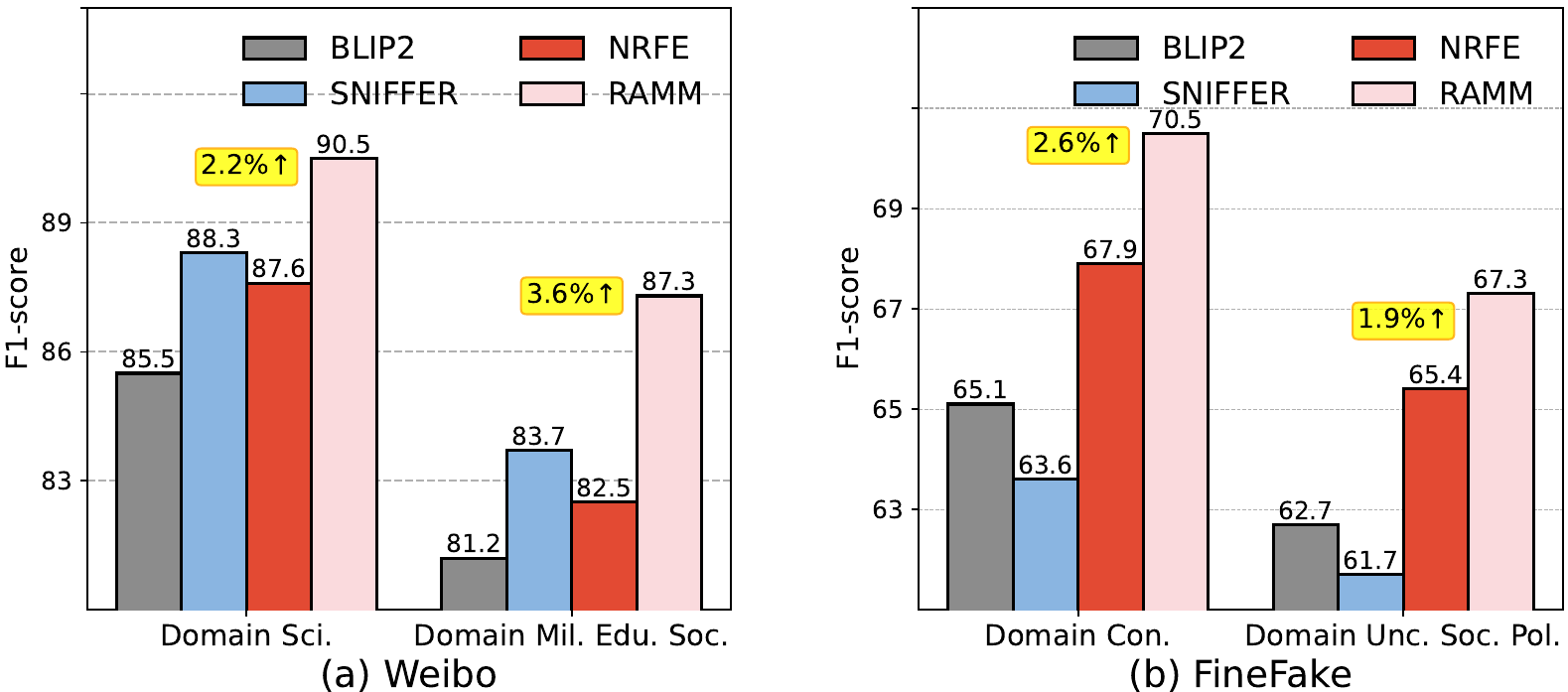}
  \caption{Performance comparison of domain generalization analysis on two datasets.}
  \label{fig:general}
  \vspace{-6pt}
\end{figure}

To verify the model's generalization performance, we conducted a Leave-domain-out experiment, which involves excluding domains during training and testing its zero-shot performance on that excluded domain. We removed one domain and three domains respectively from the Weibo dataset and the FineFake dataset for training and testing. As shown in Figure \ref{fig:general}, NRFE, SNIFFER, and BLIP2 all exhibited lower performance than RAMM after the Leave-domain-out experiment—this confirms the superior generalization of our model. This is attributed to our model’s ability to extract high-level narrative features from different domains, thereby improving the model’s associative ability and further enhancing its generalization.

\subsection{Robustness to Retrieval Noise (RQ5)}
To evaluate the robustness of our proposed model, we simulate a noisy retrieval environment by "contaminating" the top-K retrieved items for each source news. Specifically, for a given noise ratio $p$, we randomly replace $p\%$ of the top-ranked retrieved items with randomly selected  items from the  the dataset. These distractors are irrelevant to the source news. The baseline performance is established at a 0\% noise ratio. We then incrementally increase the noise ratio to 20\%, 40\%, 60\%.

As show in Table \ref{tab:robustness_all_metrics}, the robustness of our model is demonstrated by its sustained performance under noisy conditions. With minor noise interference (e.g., 20\%), the model's performance remains relatively stable, highlighting its capacity to filter secondary irrelevancies. Crucially, even under significant noise, its performance does not fall below that of leading baselines, confirming the resilient nature of our approach.

\begin{table}[t]
\centering
\small
\caption{RAMM robustness comparison on three datasets under increasing retrieval noise.}
\label{tab:robustness_all_metrics}

\setlength{\tabcolsep}{3pt} 

\begin{tabular}{@{}l ccc ccc ccc@{}}

\toprule
\multirow{2}{*}{\textbf{Noise}} & \multicolumn{3}{c}{Weibo} & \multicolumn{3}{c}{Weibo-21} & \multicolumn{3}{c}{FineFake} \\
\cmidrule(lr){2-4} \cmidrule(lr){5-7} \cmidrule(l){8-10}
 & Acc & F1 & AUC & Acc & F1 & AUC & Acc & F1 & AUC \\ 
\midrule
-20\% & 95.6 & 95.6 & 98.8 & 96.7 & 96.6 & 99.1 & 80.9 & 80.6 & 88.9 \\
-40\% & 95.2 & 95.1 & 98.1 & 96.2 & 96.2 & 98.9 & 80.3 & 80.1 & 88.6 \\
-60\% & 94.5 & 94.4 & 98.2 & 95.5 & 95.5 & 98.6 & 79.7 & 79.4 & 88.1 \\
\midrule 
\textbf{-Base(0\%)} & \textbf{96.1} & \textbf{95.9} & \textbf{98.9} & \textbf{96.9} & \textbf{96.8} & \textbf{99.1} & \textbf{81.1} & \textbf{80.9} & \textbf{89.2} \\
\bottomrule
\end{tabular}

\end{table}
\subsection{Visualization Analysis (RQ6)}
\begin{figure}[t]
  \centering
  \includegraphics[width=0.9\linewidth]{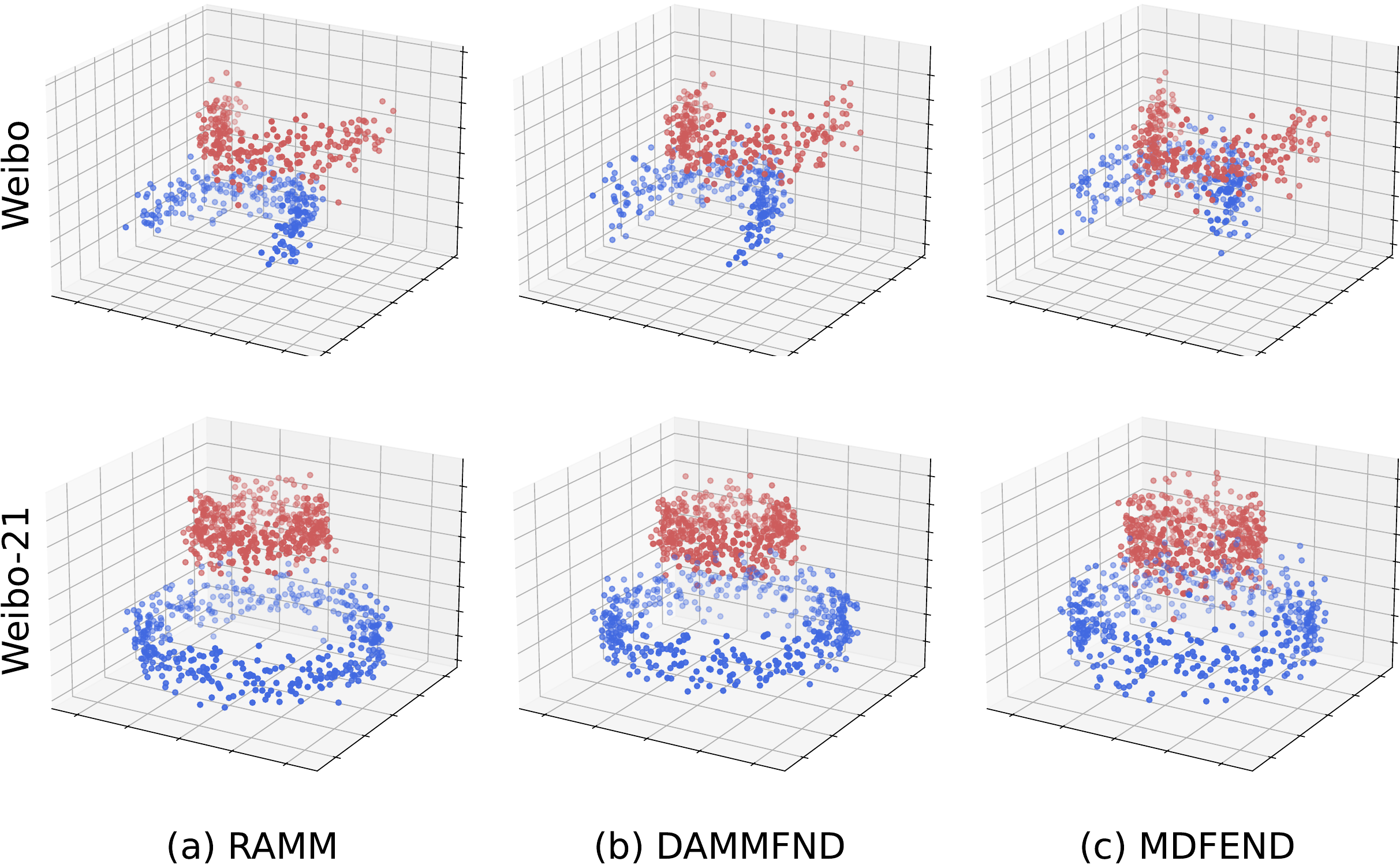}
  \caption{T-SNE showcase the classification outcomes in Weibo and Weibo-21 dataset.}
  \label{fig:tsne}
  \vspace{-6pt}
\end{figure}

Figure \ref{fig:tsne} presents t-SNE visualizations of features learned by RAMM, DAMMFND, and MDFEND on Weibo and Weibo21 test sets. RAMM generates fewer outliers in fake news representations and demonstrates reduced overlap between real and fake news embeddings compared to DAMMFND and MDFEND. These findings further substantiate RAMM’s superior performance in multimodal fake news detection.

\subsection{Robustness to LLM backbone}
\begin{figure}[h]
  \centering
  \includegraphics[width=1\linewidth]{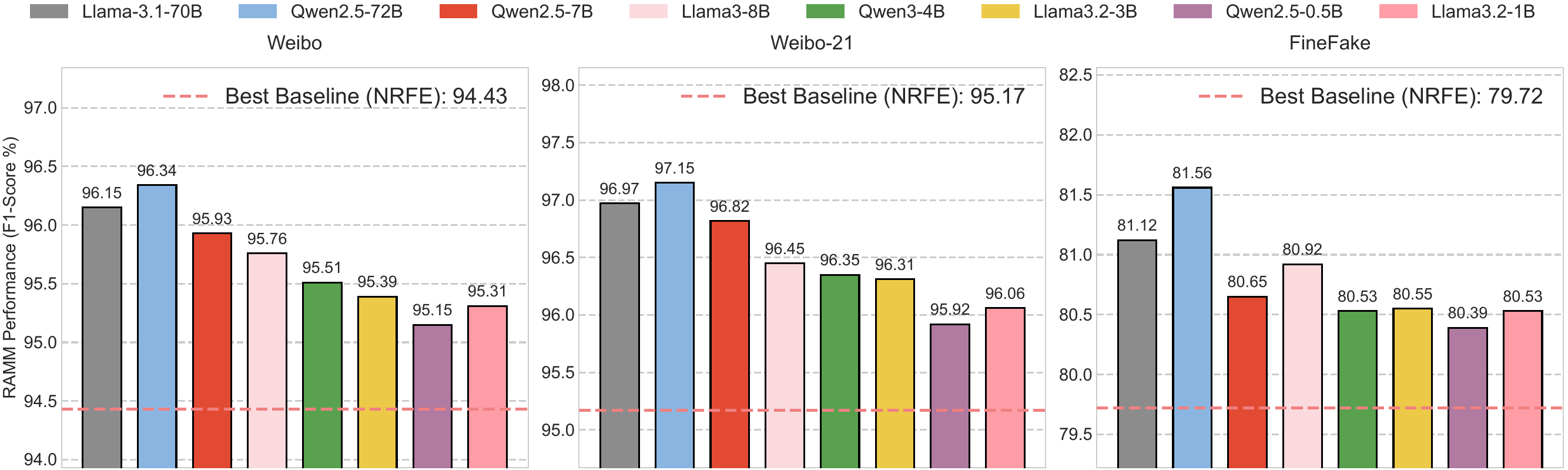}
  \caption{Performance comparison of LLM backbones.}
  \label{fig:llml}
  \vspace{-6pt}
\end{figure}

To validate the robustness of the proposed model across different backbones, extensive experiments were conducted. As illustrated in Figure \ref{fig:llml}, every configuration of our framework, regardless of the underlying LLM backbone, significantly surpasses the performance of the state-of-the-art baseline, NRFE, on the Weibo, Weibo-21, and FineFake benchmarks. Notably, Qwen2.5-72B emerges as the top-performing backbone. Due to its parameter size and reasoning complexity, we chose Qwen2.5-7B for Chinese datasets and Llama3-8B for the English dataset. These configurations achieved peak F1-scores of 95.93 (Weibo), 96.82 (Weibo-21), and 80.92 (FineFake). More importantly, the performance variance among the other leading LLMs is marginal, indicating that our framework is not merely dependent on a single model's capabilities. This high degree of performance consistency demonstrates the framework's strong robustness and its ability to effectively harness the reasoning power of diverse LLMs for reliable fake news detection.
\section{Conclusion}

In this paper, we address the cluster propagation issue in multi-domain and multi-modal fake news detection, which arises from the inability of previous models to capture the high-order narrative information underlying clusters. RAMM is designed to emulate the human fact-checking process of "association and analogy." It achieves this paradigm shift through two core components: 1) Abstract Narrative Alignment Module, which, for the first time, explicitly models high-level, cross-instance narrative consistency; 2) Semantic Representation Alignment Module, which introduces an instance-based analogical reasoning mechanism. Extensive experimental results on three
public datasets validate the efficacy of our proposed approach.

\bibliographystyle{ACM-Reference-Format}
\bibliography{main}
\clearpage
\appendix

\section{Experimental Settings}
\subsection{Datasets}
Our model is evaluated on three real-world datasets: two Chinese multi-domain datasets, Weibo \cite{wang2018eann} and Weibo-21 \cite{nan2021mdfend}, and one English multi-domain dataset, Fine-Fake \cite{zhou2024finefake}. The Weibo and Weibo-21 datasets are divided into nine domains, while FineFake contains six domains and one uncategorized domain. 
\begin{table}[htbp]
    
    \centering
    \caption{Data Statistics of Weibo}
    \label{weibo}
    \resizebox{\columnwidth}{!}{
    \begin{tabular}{lccccc}
        \toprule 
        \textbf{domain} & \textbf{Science} &\textbf{Military}  & \textbf{Education}  & \textbf{International} & \textbf{Politics}  \\
        \midrule 
        \textbf{real} & 139 & 160 & 273 & 45 & 112  \\
        \textbf{fake} & 91 & 122 & 148 & 41 & 164  \\
        \midrule
        \textbf{all} & 230 & 282 & 421 & 86 & 276  \\
        \midrule
        \midrule
        \textbf{domain} & \textbf{Health} & \textbf{Finance} & \textbf{Entertainment} & \textbf{Society} & \textbf{All}  \\
        \midrule
        \textbf{real} & 186& 137& 1,120 &1,443& 3,615 \\
        \textbf{fake} & 519 &143 &508 &2,372 &4,108 \\
        \midrule
        \textbf{all} & 705& 280 &1,628 &3,815 &7,723 \\
        \bottomrule 
    \end{tabular}}
\end{table}

\begin{table}[h]
    
    \centering
    \caption{Data Statistics of Weibo21}
    \label{weibo21}
    \resizebox{\columnwidth}{!}{
    \begin{tabular}{lccccc}
        \toprule 
        \textbf{domain} & \textbf{Science} &\textbf{Military}  & \textbf{Education}  & \textbf{Disasters} & \textbf{Politics}  \\
        \midrule 
        \textbf{real} & 143 & 121 & 243 & 185 & 306  \\
        \textbf{fake} & 93 & 222 & 248 & 591 & 546  \\
        \midrule
        \textbf{all} & 236 & 343 & 491 & 776 & 852  \\
        \midrule
        \midrule
        \textbf{domain} & \textbf{Health} & \textbf{Finance} & \textbf{Entertainment} & \textbf{Society} & \textbf{All}  \\
        \midrule
        \textbf{real} & 485& 959& 1,000 &1,198& 4,640 \\
        \textbf{fake} & 515 &362 &440 &1,471 &4,488 \\
        \midrule
        \textbf{all} & 1,000& 1,321 &1,440 &2,669 &9,128 \\
        \bottomrule 
    \end{tabular}}
\end{table}

\begin{table}[h]
    
    \centering
    \caption{Data Statistics of FineFake}
    \label{finefake}
    \resizebox{\columnwidth}{!}{
    \begin{tabular}{lccccc}
        \toprule 
        \textbf{domain} & \textbf{Politics} &\textbf{Entertainment}  & \textbf{Business }  & \textbf{Health} & \textbf{Society}  \\
        \midrule 
        \textbf{real} & 3,722 & 2,514 & 527 & 438 & 2,236  \\
        \textbf{fake} & 2,005 & 1,185 & 476 & 272 & 1,703  \\
        \midrule
        \textbf{all} & 5,727 & 3,699 & 1,003 & 710 & 3,939  \\
        \midrule
        \midrule
        \textbf{domain} & \textbf{Conflict} & \textbf{Stream } & \textbf{Official } & \textbf{Fact-Check} & \textbf{All}  \\
        \midrule
        \textbf{real} & 979& 3,895& 4,138 &2,474& 10,507 \\
        \textbf{fake} & 739 &1,105 &215 &5,082 &6,402 \\
        \midrule
        \textbf{all} & 1,718& 5,000 &4,353 &7,556 &16,909 \\
        \bottomrule 
    \end{tabular}}
\end{table}
\subsection{Baselines}
To comprehensively assess the performance of our proposed model, we compare it against baseline methods organized into three categories: (1) single-modal multi-domain methods, including: \textbf{MDFEND} \cite{nan2021mdfend}, \textbf{M$^3$DFEND} \cite{zhu2022memory} and \textbf{PLDFEND} \cite{peng2023prompt}; (2) multi-modal multi-domain methods, including: \textbf{KATMF} \cite{song2021knowledge}, \textbf{MMDFND} \cite{tong2024mmdfnd}, \textbf{DAMMFND} \cite{lu2025dammfnd}, \textbf{IMOL} \cite{zeng2025imol} and \textbf{MMHT} \cite{yang2025macro}; (3) MLLM methods, including: \textbf{BLIP2} \cite{li2023blip}, \textbf{SNIFFER} \cite{qi2024sniffer}, \textbf{GSFND} \cite{tong2025generate}, \textbf{NRFE} \cite{zhang2025llms}. Descriptions of baseline methods are as follows:

\begin{enumerate}
    \item \textbf{Single-Modal Multi-Domain Methods}:
    \begin{itemize}
        \item \textbf{MDFEND}, which utilizes a domain gate to aggregate multiple representations extracted by a mixture of experts.
        \item \textbf{M$^3$DFEND}, which models news from semantic, emotional, and stylistic perspectives, and uses a domain memory bank and domain adapter to address domain shift and incomplete domain labels.
        \item \textbf{PLDFEND}, which applies prompt-based learning to multi-domain fake news detection by designing sophisticated prompt templates and corresponding verbalizers.
    \end{itemize}

    \item \textbf{Multimodal Multi-Domain Methods}:
    \begin{itemize}
        \item \textbf{KATMF}, which employs adversarial multi-task learning and a knowledge-enhanced Transformer to capture the differences in feature distributions across various news domains.
        \item \textbf{MMDFND}, which utilizes an improved PLE (Progressive Layered Extraction) module to capture both the commonalities and specificities among domains and incorporates AdaIN (Adaptive Instance Normalization) for multimodal fusion.
        \item \textbf{DAMMFND}, which extracts more accurate domain information through Domain Disentanglement, while simultaneously mitigating negative transfer between domains.
        \item \textbf{IMOL}, which addresses the issue of incomplete modalities in multi-domain fake news video detection by reconstructing missing modalities through cross-modal consistency and refining transferable knowledge via cross-sample associative reasoning.
        \item \textbf{MMHT}, which utilizes a macro-micro hierarchical Transformer architecture for multimodal fake news detection.
    \end{itemize}

    \item \textbf{MLLM Methods}:
    \begin{itemize}
        \item \textbf{GSFND}, adopting an LLM to generate fake news in three different styles which are later incorporated into the training set, to augment the representation of fake news.
        \item \textbf{SNIFFER}, a multimodal large language model designed for misinformation detection and explanation, which employs a two-stage instruction tuning process on InstructBLIP.
        \item \textbf{NRFE}, which leverages positive or negative news-reasoning pairs for learning the semantic consistency between them.
        \item \textbf{BLIP2}, which bootstraps vision-language pre-training from frozen pre-trained image encoders and large language models using a lightweight Querying Transformer to bridge the modality gap.
    \end{itemize}
\end{enumerate}
\subsection{Implementation Details}
\label{sec:imple}
To extract a stable and high-quality core narrative summary, $\mathcal{A}_{u}$, from the prompt $\mathcal{P}_{u}$,  we utilize the powerful \texttt{Qwen3-MAX} Vision-Language Model. The choice is deliberate: it allows for the direct, joint processing of both text and image modalities, thereby avoiding the information loss that can occur when relying on intermediate image captions or descriptions. 

The prompt template designed for this narrative extraction task is as follows:
\begin{tcolorbox}[
    colback=black!5!white,
    colframe=black!75!black,
    title=\textbf{Prompt for Multimodal Narrative Distillation},
    fonttitle=\bfseries,
    arc=2mm,
    boxrule=1pt
]
\textbf{SYSTEM PROMPT}
\begin{itemize}[leftmargin=*, topsep=0pt, itemsep=0pt, partopsep=0pt, parsep=2pt]
    \item \textbf{Persona:} You are an expert AI assistant specializing in media analysis and misinformation detection.
    \item \textbf{Core Task:} Your task is to analyze multimodal news content (text and image) and distill its core abstract narrative or central claim into a single, concise sentence.
    \item \textbf{Constraints:}
    \begin{itemize}[leftmargin=*, topsep=0pt, itemsep=0pt]
        \item Focus strictly on the underlying message, not superficial details.
        \item Remain objective and neutral; do not judge the veracity of the claim.
        \item The output \textbf{MUST} be a single sentence.
        \item Your response must contain \textbf{ONLY} the narrative sentence, with no preamble, explanations.
    \end{itemize}
\end{itemize}
\hrule
\vspace{0.5em}
\textbf{USER PROMPT}

\vspace{0.2em}
\textbf{Example 1:}
\begin{itemize}[leftmargin=*, topsep=0pt, itemsep=0pt, partopsep=0pt, parsep=2pt]
    \item \textbf{Text:} \textit{"Explosive footage! A poll worker in XX County was caught on a hidden camera shredding pro-government ballots to rig the election. The proof is undeniable!"}
    \item \textbf{Image:} \texttt{[Multimodal Input: A blurry image of a person standing near paper bins.]}
\end{itemize}
\textbf{Core Narrative:} \textit{Election fraud has been proven beyond doubt.}

\vspace{0.2em}
\textbf{Example 2:}
\begin{itemize}[leftmargin=*, topsep=0pt, itemsep=0pt, partopsep=0pt, parsep=2pt]
    \item \textbf{Text:} \textit{"Data never lies! This 90-degree voting curve is statistically impossible unless there is some algorithm at work to cheat!"}
    \item \textbf{Image:} \texttt{[Multimodal Input: A line graph of vote counts showing a sudden, sharp vertical spike.]}
\end{itemize}
\textbf{Core Narrative:} \textit{Election fraud has been proven beyond doubt.}

\vspace{0.5em}
\hrule
\vspace{0.5em}

\textbf{Final Query:}
\begin{itemize}[leftmargin=*, topsep=0pt, itemsep=0pt, partopsep=0pt, parsep=2pt]
    \item \textbf{Text:} \texttt{\{text\_content\}}
    \item \textbf{Image:} \texttt{[Multimodal Input: News image.]}
\end{itemize}
\textbf{Core Narrative:}
\end{tcolorbox}

For each sample in our datasets, we construct a multimodal prompt where the model is presented with both the original news text and its associated image. During the generation process, we set the decoding \texttt{temperature} to 0.7 to strike a balance between creativity and factual consistency in the generated summaries.

The objective of this pre-processing step is to distill the raw, diverse multimodal content into a focused and standardized textual narrative, which is crucial for the subsequent narrative embedding and retrieval stages. 
This process can be formally expressed as:
\begin{equation}
    \mathcal{A}_{u} = \text{Qwen3-MAX}(\mathcal{P}_{u})
    \label{eq:narrative_extraction}
\end{equation}

\section{Theoretical Analysis of Common Information Bottleneck Loss}
\label{sec:theory}
We posit that a robust representation for fake news detection should $(a)$ encapsulate the core, invariant narrative across related instances; $(b)$ preserve the unique, critical information of the specific instance;  $(c)$ remain concise to prevent overfitting. We demonstrate that conventional objectives fail to satisfy these criteria simultaneously, whereas our proposed Common Information Bottleneck Loss (CIBL) provides a principled framework to achieve this balance.

Let $\mathbf{h}_u \in \mathcal{H}$ be the representation of a news sample $u$ and $\mathbf{h}_u^+ \in \mathcal{H}$ be its synthesized positive counterpart. The objective is to learn a latent representation $\mathbf{z}_u \in \mathcal{Z}$ via an encoder $q_\phi(\mathbf{z}_u | \mathbf{h}_u, \mathbf{h}_u^+)$.

\paragraph{1. Limitations of Standard Contrastive Learning (CL)}
The standard contrastive objective, such as InfoNCE, maximizes a variational lower bound on the mutual information (MI) between the encoded views:
\begin{equation}
\mathcal{L}_{\text{CL}} \propto -I(f_\theta(\mathbf{h}_u); f_\theta(\mathbf{h}_u^+))
\end{equation}
where $f_\theta$ is the encoder. Let the information content of the representations be decomposed as $\mathbf{h}_u \triangleq C \cup U$ and $\mathbf{h}_u^+ \triangleq C \cup U^+$, where $C$ is the common narrative and $U, U^+$ are unique components. The CL objective implicitly encourages the encoder to become invariant to non-common parts, leading to:
\begin{equation}
f_\theta(\mathbf{h}_u) \approx f_\theta(C) \implies I(f_\theta(\mathbf{h}_u); U) \to 0
\end{equation}
This phenomenon, known as representation collapse, is detrimental if the unique information $U$ is predictive of the ground-truth veracity label $y$, i.e., if $I(y; U | C) > 0$. The learned representation becomes suboptimal, as valuable information is discarded.

\paragraph{2. Limitations of the Canonical Information Bottleneck (IB)}
The canonical IB principle \citep{tishby2000information} is formulated as:
\begin{equation}
\max_{\mathbf{z}} I(\mathbf{z}; \mathbf{y}) - \beta I(\mathbf{z}; \mathbf{x})
\end{equation}
Adapting this to our self-supervised setting by setting $\mathbf{x} \equiv \mathbf{h}_u$ and $\mathbf{y} \equiv \mathbf{h}_u^+$, the objective becomes:
\begin{equation}
\max_{\mathbf{z}_u} I(\mathbf{z}_u; \mathbf{h}_u^+) - \beta I(\mathbf{z}_u; \mathbf{h}_u)
\end{equation}
This formulation is inherently flawed. 

The compression term, $-\beta I(\mathbf{z}_u; \mathbf{h}_u)$, penalizes any information retained from $\mathbf{h}_u$, including the common information $C$ which is essential for maximizing the predictive term $I(\mathbf{z}_u; \mathbf{h}_u^+)$. The objective creates a conflict rather than encouraging the desired disentanglement of common and unique information.

\paragraph{3. The CIBL as a Variational Information-Theoretic Objective}
CIBL resolves the aforementioned limitations by jointly optimizing three information-theoretic desiderata. The full objective is a variational upper bound on the following functional:
\begin{align}
    \mathcal{F}(\phi, \psi) &= \underbrace{-\lambda_1 I(\mathbf{z}_u; \mathbf{h}_u^+)}_{\text{Alignment}} \nonumber  + \underbrace{(- \lambda_2 \mathbb{E}_{q_\phi}[\log p_\psi(\mathbf{h}_u|\mathbf{z}_u)])}_{\text{Reconstruction}} \\
    &\quad+ \underbrace{\lambda_3 \text{KL}(q_\phi(\mathbf{z}_u|\mathbf{h}_u, \mathbf{h}_u^+)\|p(\mathbf{z}))}_{\text{Compression}}
\label{eq:cibl_full}
\end{align}
Each component of our loss, $\mathcal{L}_{\text{CIBL}} = \lambda_1\mathcal{L}_{\text{align}} + \lambda_2 \mathcal{L}_{\text{recon}} + \lambda_3 \mathcal{L}_{\text{compress}}$, corresponds to a bound on these terms.

\begin{itemize}
    \item \textbf{Alignment:} $\mathcal{L}_{\text{align}}$ is the InfoNCE loss, which maximizes a lower bound on MI \citep{oord2018representation, poole2019variational}:
    \begin{equation}
    -\mathcal{L}_{\text{align}} \leq I(\mathbf{z}_u; \mathbf{h}_u^+)
    \end{equation}
    This forces $\mathbf{z}_u$ to be maximally predictive of the common abstract narrative contained within $\mathbf{h}_u^+$.

    \item \textbf{Reconstruction:} $\mathcal{L}_{\text{recon}}$ is the negative log-likelihood of a decoder $p_\psi$. 
    
    For a Gaussian decoder, $p_\psi(\mathbf{h}_u|\mathbf{z}_u) = \mathcal{N}(g_\psi(\mathbf{z}_u), \sigma^2 I)$, this reduces to the mean squared error $\|\mathbf{h}_u - g_\psi(\mathbf{z}_u)\|_2^2$. This term maximizes a lower bound on the MI required to reconstruct the original input:
    \begin{equation}
    I(\mathbf{z}_u; \mathbf{h}_u) \geq \mathcal{H}(\mathbf{h}_u) - \mathbb{E}_{\mathbf{z}_u \sim q_\phi}[-\log p_\psi(\mathbf{h}_u|\mathbf{z}_u)] = \mathcal{H}(\mathbf{h}_u) - \mathcal{L}_{\text{recon}}
    \end{equation}
    where $\mathcal{H}(\cdot)$ is the differential entropy. Minimizing $\mathcal{L}_{\text{recon}}$ prevents the loss of unique, essential information from $\mathbf{h}_u$, directly addressing the limitation of standard CL.

    \item \textbf{Compression:} $\mathcal{L}_{\text{compress}}$ is the Kullback-Leibler divergence between the posterior and a prior, which serves as a variational upper bound on the MI between the latent code and the complete input pair:
    \begin{equation}
    \mathcal{L}_{\text{compress}} = \text{KL}(q_\phi(\mathbf{z}_u|\mathbf{h}_u, \mathbf{h}_u^+)\|p(\mathbf{z})) \geq I(\mathbf{z}_u; \mathbf{h}_u, \mathbf{h}_u^+)
    \end{equation}
    This term regularizes the complexity of the learned representation $\mathbf{z}_u$, promoting conciseness.
\end{itemize}

\paragraph{4. Generalization Bound Perspective}
The superiority of CIBL is further substantiated by learning theory. From a PAC-Bayesian standpoint, the generalization error $\mathcal{E}_{\text{gen}}$ is bounded by the empirical error $\mathcal{E}_{\text{emp}}$ plus a complexity term related to the KL divergence between the posterior and prior distributions over model parameters \citep{xu2017information}. This complexity term can be linked to the MI between the learned representation and the training data $\mathcal{D}_n$:
\begin{equation}
\mathcal{E}_{\text{gen}} \le \mathcal{E}_{\text{emp}} + \mathcal{O}\left(\sqrt{\frac{I(\mathbf{z}; \mathcal{D}_n)}{n}}\right)
\end{equation}
Standard CL lacks explicit control over $I(\mathbf{z}; \mathcal{D}_n)$, potentially leading to a loose bound and overfitting. In contrast, CIBL, through the $\mathcal{L}_{\text{compress}}$ term, directly minimizes an upper bound of this MI. By jointly minimizing $\mathcal{L}_{\text{align}}$ and $\mathcal{L}_{\text{recon}}$ (to reduce $\mathcal{E}_{\text{emp}}$) and $\mathcal{L}_{\text{compress}}$ (to tighten the bound), CIBL learns a minimally sufficient representation that is provably more likely to generalize well.

In summary, CIBL provides a more comprehensive learning objective that forces the model to disentangle and preserve information in a structured manner, directly yielding representations with a tighter information-theoretic generalization bound compared to conventional methods.
\end{document}